\documentclass[runningheads]{llncs}

\usepackage[T1]{fontenc}
\usepackage{graphicx}
\usepackage{calc}
\usepackage{amssymb}
\usepackage{amsmath}
\usepackage{doi}
\usepackage{tikz}
\usetikzlibrary{shapes.geometric, arrows.meta, positioning, calc, shadows, fit, backgrounds}
\usepackage{float}       
\usepackage{pgfplots}
\usepackage{hyperref}
\usepackage{cite}
\hypersetup{
    hidelinks,
    colorlinks=false,
    pdfborder={0 0 0}
}
\pgfplotsset{compat=1.17}

\tikzset{
    process/.style={
        rectangle, 
        draw=black!70, 
        fill=blue!5, 
        thick, 
        minimum width=2.5cm, 
        minimum height=1cm, 
        align=center, 
        rounded corners=3pt,
        drop shadow,
        font=\sffamily\footnotesize
    },
    model/.style={
        rectangle, 
        draw=violet!70, 
        fill=violet!10, 
        thick, 
        minimum width=2.2cm, 
        minimum height=0.8cm, 
        align=center, 
        rounded corners=3pt,
        drop shadow,
        font=\sffamily\scriptsize
    },
    data/.style={
        trapezium, 
        trapezium left angle=70, 
        trapezium right angle=110, 
        draw=orange!70, 
        fill=orange!10, 
        thick, 
        minimum width=2cm, 
        minimum height=0.8cm, 
        align=center,
        font=\sffamily\footnotesize
    },
    decision/.style={
        diamond, 
        draw=red!70, 
        fill=red!10, 
        thick, 
        aspect=2, 
        align=center, 
        inner sep=2pt,
        font=\sffamily\footnotesize
    },
    arrow/.style={
        -{Latex[length=3mm]}, 
        thick, 
        draw=gray!70
    },
    group/.style={
        draw=gray!40, 
        dashed, 
        inner sep=10pt, 
        rounded corners=5pt, 
        fill=gray!5
    }
}

\begin{document}

\title{An Agentic Evaluation Architecture for Historical Bias Detection in Educational Textbooks}
\titlerunning{Agentic Bias Detection in Educational Textbooks}

\author{Gabriel Ștefan\orcidID{0009-0007-4193-7181} \and
Adrian Marius Dumitran\orcidID{0009-0005-3547-5772}}

\authorrunning{G. Ștefan and A. M. Dumitran}

\institute{University of Bucharest, Bucharest, Romania \\
\email{gabrielstefan04@gmail.com, marius.dumitran@unibuc.ro}}

\maketitle              

\begin{abstract}
History textbooks often contain implicit biases, nationalist framing, and selective omissions that are difficult to audit at scale. We propose an agentic evaluation architecture comprising a multimodal screening agent, a heterogeneous jury of five evaluative agents, and a meta-agent for verdict synthesis and human escalation. A central contribution is a Source Attribution Protocol that distinguishes textbook narrative from quoted historical sources, preventing the misattribution that causes systematic false positives in single-model evaluators.

In an empirical study on Romanian upper-secondary history textbooks, 83.3\% of 270 screened excerpts were classified as pedagogically acceptable (mean severity 2.9/7), versus 5.4/7 under a zero-shot baseline, demonstrating that agentic deliberation mitigates over-penalization. In a blind human evaluation (18 evaluators, 54 comparisons), the Independent Deliberation configuration was preferred in 64.8\% of cases over both a heuristic variant and the zero-shot baseline. At approximately \$2 per textbook, these results position agentic evaluation architectures as economically viable decision-support tools for educational governance.
\end{abstract}


\section{Introduction}
\label{sec:introduction}

Educational materials are curated narratives that shape collective memory,
civic identity, and historical interpretation~\cite{apple2014official,pingel2010unesco}. National curriculum reforms introduce large volumes of
new instructional materials simultaneously, creating an auditing demand
that manual expert review cannot satisfy at scale.

Deploying LLMs as curriculum auditing tools presents two domain-specific
failure modes. First, safety-aligned models frequently misclassify
historically sensitive but factually accurate content as
problematic~\cite{rottger2023xstest}. Second, LLMs fail to distinguish
authorial narrative from quoted primary sources, producing spurious bias
attributions. Furthermore, a single-model zero-shot baseline suffers from long-context attention degradation~\cite{liu2024lost,hsieh2024ruler}, extracting only a fraction of problematic passages and inflating severity to a mean of 5.4/7 with no excerpt rated below 4.

To address these failure modes, we propose an \textit{agentic evaluation
architecture}: a screening agent for broad-coverage discovery, a
heterogeneous jury for deliberative severity assessment, and a meta-agent
for synthesis and human escalation~\cite{du2023improving}. At approximately
\$2 per textbook, the system surfaces prioritized controversies before
materials are published, without rendering autonomous final judgments.

This paper makes the following contributions:
\begin{enumerate}
    \item An agentic evaluation architecture that reduces mean severity
    from 5.4/7 (zero-shot) to 2.9/7, classifying 83.3\% of screened
    excerpts as pedagogically acceptable while producing structured
    outputs suitable for editorial governance workflows.

    \item A Source Attribution Protocol that enforces explicit
    classification of excerpts as authorial narrative or quoted
    historical material, reducing false positives from primary source
    misattribution.

    \item An empirical case study on publicly available Romanian
    upper-secondary history textbooks, demonstrating that the
    architecture surfaces historiographically significant patterns
    at a scale and consistency that manual review cannot match.
\end{enumerate}

\section{Related Work}
\label{sec:related_work}

\subsection{Computational Curriculum Analysis}
Textbook analysis has traditionally relied on manual, qualitative examination 
of national narratives and historical framing~\cite{apple2014official,pingel2010unesco}. 
Early computational work applied lexicon-based methods to quantify representational 
patterns~\cite{lucy2020content}, while more recent approaches use neural models 
for readability and framing analysis~\cite{zhai2021review}. Both 
generations of methods operate on plain-text extractions, discarding the spatial 
structure---sidebars, marginalia, captions---essential for interpreting complex 
pedagogical layouts~\cite{xu2020layoutlm,bai2023qwenvl}. Our work addresses 
this gap by combining layout-aware multimodal screening with deliberative 
adjudication.

\subsection{Attribution Errors and Safety Inflation in LLM Evaluators}
Attribution errors, where models conflate cited evidence with authorial 
claims, are a documented limitation of LLM evaluators~\cite{ji2023survey}. 
Existing mitigations via retrieval augmentation~\cite{lewis2020retrieval} or 
post-hoc verification~\cite{zhang2025siren} target generation settings and do 
not transfer to evaluative tasks assessing existing material. Standard frameworks 
such as G-Eval~\cite{liu2023gevalnlgevaluationusing} and 
MT-Bench~\cite{zheng2024judging} lack explicit mechanisms to distinguish endorsed 
narrative from quoted historical sources, resulting in systematic 
over-penalization of factually accurate content~\cite{rottger2023xstest}. We 
address both failure modes through a Source Attribution Protocol that enforces 
this distinction as a constrained intermediate representation prior to evaluation.

\subsection{Multi-Agent Systems and AI in Education}
Multi-agent architectures have demonstrated superior robustness and factuality 
over single-model pipelines in reasoning-intensive 
tasks~\cite{du2023improving,chan2023chateval,chen2023agentverse,wu2023autogen}. In AI in Education, 
LLMs have been applied to tutoring and content 
creation~\cite{kasneci2023chatgpt}, but their application 
to curriculum governance remains limited~\cite{unesco2023guidance,luckin2016intelligence}. 
We bridge these domains by adapting deliberative multi-agent evaluation---
traditionally used to improve generative quality---into a conservative reliability 
filter for large-scale educational auditing, where the objective is to suppress 
weakly supported evaluative claims rather than enhance generation. 

\section{Proposed Pipeline}
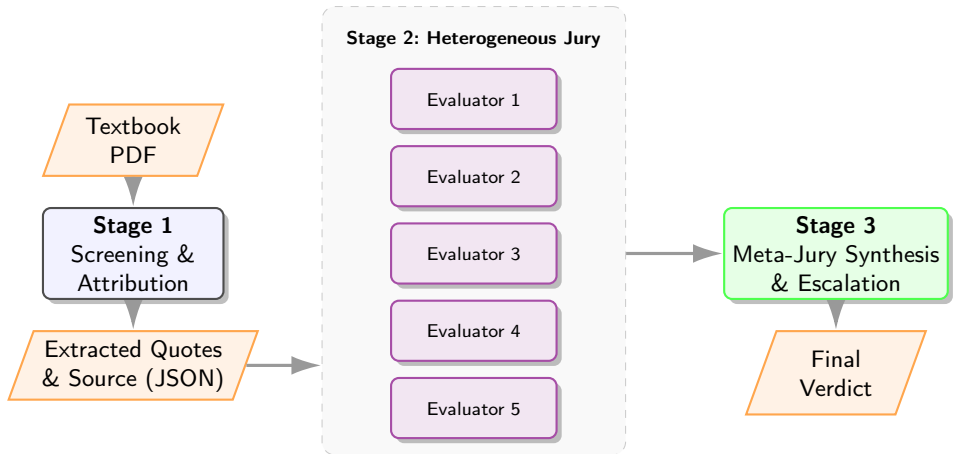
\begin{figure}[H]
\centering
\pgfdeclarelayer{background}
\pgfdeclarelayer{foreground}
\pgfsetlayers{background,main,foreground}
\begin{tikzpicture}[
    node distance=0.4cm,
    font=\sffamily\scriptsize,
    arrow/.style={-{Latex[length=4mm, width=2.5mm]}, line width=1.2pt, draw=gray!80}
]

\node[model] (eval3) {Evaluator 3};
\node[model, above=0.2cm of eval3] (eval2) {Evaluator 2};
\node[model, above=0.2cm of eval2] (eval1) {Evaluator 1};
\node[model, below=0.2cm of eval3] (eval4) {Evaluator 4};
\node[model, below=0.2cm of eval4] (eval5) {Evaluator 5};

\node[font=\sffamily\scriptsize\bfseries, above=0.15cm of eval1] (jury_title) {Stage 2: Heterogeneous Jury};

\begin{pgfonlayer}{background}
    \node[group, fit=(jury_title)(eval1)(eval5), inner sep=6pt] (jury_box) {};
\end{pgfonlayer}

\node[process, left=2.2cm of eval3, minimum width=2.4cm, align=center] (screening) {\textbf{Stage 1}\\Screening \&\\Attribution};
\node[data, above=0.4cm of screening, minimum width=2.4cm, align=center] (input) {Textbook\\PDF};
\node[data, below=0.4cm of screening, minimum width=2.4cm, align=center] (candidates) {Extracted Quotes\\\& Source (JSON)};

\node[process, right=2.2cm of eval3, minimum width=2.4cm, align=center, fill=green!10, draw=green!70] (metajury) {\textbf{Stage 3}\\Meta-Jury Synthesis\\\& Escalation};
\node[data, below=0.4cm of metajury, minimum width=2.4cm, align=center] (output) {Final\\Verdict};

\draw[arrow] (input) -- (screening);
\draw[arrow] (screening) -- (candidates);

\draw[arrow] (candidates.east) -- (jury_box.west |- candidates.east);

\draw[arrow] (jury_box.east |- metajury.west) -- (metajury.west);

\draw[arrow] (metajury) -- (output);

\end{tikzpicture}
\caption{The three-stage agentic evaluation architecture. The flow illustrates the screening agent, heterogeneous jury, and meta-agent for verdict synthesis.}
\label{fig:pipeline_diagram}
\end{figure}

The proposed system instantiates an \textit{agentic evaluation
architecture} for curriculum auditing, in which specialized agents with
distinct roles collaborate to produce auditable, evidence-backed verdicts.
Unlike single-pass LLM classification, where a single model is prompted
to both identify and assess bias in one step, the architecture
decomposes the task across three agent layers: a screening agent
performing broad-coverage issue discovery, a jury of heterogeneous
evaluative agents conducting independent severity assessment, and a
meta-agent synthesizing verdicts and triggering escalation to human
reviewers. This decomposition is motivated by the failure modes of monolithic
classification in deployment: over-sensitivity to historically sensitive
content, source attribution errors, and the lack of interpretable reasoning chains in single-model verdicts (see Fig.~\ref{fig:pipeline_diagram}).

\subsection{Stage 1: Broad-Coverage Screening and Joint Attribution}
\label{sec:screening}

The screening agent performs high-sensitivity discovery using a
multimodal large language model that ingests textbook pages as
high-resolution images, preserving layout information---headings,
marginalia, sidebars, and image captions---that plain-text extraction
discards. The agent is parameterized to favor sensitivity over
precision: its role is to surface any passage that may warrant
scrutiny, including nationalist framing, omission of historical
context, or uncontextualized primary sources. False positives are
tolerated by design, as all flagged excerpts are subject to jury
adjudication in Stage 2.

Rather than treating attribution as an implicit reasoning step,
the screening agent enforces a strict \textbf{Source Attribution
Schema} during extraction. The model must simultaneously flag the excerpt and classify it into one of two mutually exclusive categories: \textit{Textbook Narrative} (the authorial explanatory voice) or \textit{Primary Source Usage} (quoted historical materials). This joint extraction ensures that jury agents evaluate each excerpt
with the correct pedagogical frame: language that would trigger a
bias classification in isolation may be entirely appropriate when
recognized as a quoted historical source.

The use of a single model in Stage 1 is a deliberate design choice.
Because all flagged excerpts are subject to independent jury adjudication
in Stage 2, the screening agent functions as a high-recall filter rather
than a final classifier: false positives are tolerated and corrected
downstream, while the cost of a false negative---a biased excerpt missed
entirely---is irreversible within the pipeline. Concentrating sensitivity
in a single, well-parameterized multimodal model avoids the coordination
overhead of an ensemble while preserving the recall objective. The risk
of missed excerpts due to model-specific blind spots remains a limitation
discussed in Section~\ref{sec:limitations}.

Screening decisions are made traceable downstream through the structured 
JSON output: each flagged excerpt carries an explicit attribution label 
and a textual reasoning field, which are passed directly to Stage 2 
jurors, ensuring that the screener's classification rationale is 
inspectable at the jury stage even without access to the screening 
model's internals.

\subsection{Stage 2: Multi-Model Jury Adjudication}
\label{sec:jury}

Each excerpt flagged by the screening agent, together with its
attribution label, is independently assessed by a jury of five
heterogeneous evaluative agents with differing alignment profiles
and reasoning architectures. Each agent independently assigns a severity score on a 7-point ordinal scale anchored to pedagogical impact (see Table~\ref{tab:severity_scale}).

\begin{table}[H]
\centering
\small
\caption{The 7-point ordinal severity scale used by jury agents}
\label{tab:severity_scale}
\setlength{\tabcolsep}{4pt}
\begin{tabular}{ll}
\hline
\textbf{Score} & \textbf{Description} \\
\hline
\textbf{1 -- Neutral} & Pedagogically sound or properly contextualized source. \\
\textbf{2 -- Negligible} & Stylistic choices without substantive bias. \\
\textbf{3 -- Minor} & Lack of secondary perspective or slight tonal loading. \\
\textbf{4 -- Moderate} & Loaded language, stereotyping, or insufficient context. \\
\textbf{5 -- Significant} & Selective omission of key facts or one-sided narratives. \\
\textbf{6 -- Severe} & Nationalist myth-making, whitewashing, or propaganda. \\
\textbf{7 -- Harmful} & Hate speech, fabrication, or incitement as instruction. \\
\hline
\end{tabular}
\end{table}

The 7-point ordinal severity scale follows psychometric recommendations 
for evaluative rating tasks~\cite{joshi2015,preston2000}, with each level 
anchored to a concrete pedagogical consequence rather than a linguistic 
surface feature~\cite{pingel2010unesco}.

\subsubsection{Bias Taxonomy Classification}
In addition to a severity score, each agent classifies the excerpt
using a pre-defined taxonomy of 15 bias labels covering four
domains: \textit{Language \& Framing}, \textit{Perspective \&
Representation}, \textit{Structure \& Emphasis}, and \textit{Source
Handling}. This taxonomy provides granular diagnostic output beyond
a scalar severity score, enabling the final report to identify not
just the severity of a concern but its historiographical character.
The taxonomy was constructed by the authors to operationalize 
dimensions of historiographical bias drawn from curriculum analysis 
and history education didactics~\cite{pingel2010unesco,apple2014official,lucy2020content,stradling2003}, extended to address failure modes specific 
to post-communist historiography. The four domains reflect established 
analytical categories in textbook research, though formal inter-rater 
validation remains a direction for future work.

\subsubsection{Structured Jury Output}
To ensure consistent downstream aggregation across heterogeneous models, all jurors must return a standardized JSON object. This object contains an \texttt{attribution} flag (\textit{Textbook Narrative} or \textit{Primary Source Usage}), a taxonomy \texttt{category}, an integer \texttt{severity} score $\in [1,7]$, a \texttt{confidence} float $\in [0.0, 1.0]$, and a textual \texttt{reasoning} justification. Crucially, requiring an explicit confidence score alongside the severity rating allows the meta-agent to appropriately weight juror agreements during synthesis.

\subsubsection{Calibration Instruction.}
A deliberate design choice in the jury prompt is the inclusion of the
instruction: \textit{``You are encouraged to assign low severity or dismiss
concerns when appropriate.''} This instruction reflects a conscious calibration
decision: given that the screening stage is parameterized for high sensitivity
(Section~\ref{sec:screening}), the jury stage must compensate by applying
a conservative severity prior. 
It should be noted, however, that this instruction is a confounding variable
in the comparison between the multi-agent pipeline and the zero-shot baseline:
the two configurations differ not only in architecture (single vs. multi-agent)
but also in explicit severity prior. Future work should isolate these variables
through ablation (see Section~\ref{sec:limitations}). All system prompts used across pipeline stages are available in the supplementary repository,\footnote{\url{https://github.com/submission-its/bias-detection}} alongside the implementation notebooks. All stages employ zero-shot prompting with no in-context examples.

\subsection{Stage 3: Meta-Jury Aggregation and Escalation}
\label{sec:meta}

The meta-agent receives the complete output tuple from each jury
agent (category, severity, confidence, and reasoning) and
synthesizes a single final verdict. To evaluate the optimal
resolution strategy for inter-agent disagreements, we implement
and compare two distinct meta-agent prompting configurations:

\begin{enumerate}
    \item \textbf{Heuristic Aggregation Strategy:} In this configuration, the Meta-Jury is strictly prompted to follow a mathematical decision tree. It adopts consensus only if high-confidence jurors ($confidence > 0.7$) agree, computes confidence-weighted averages for minor disagreements, and automatically flags the case for human review if juror severity scores diverge by more than 1.5 points.
    
    \item \textbf{Independent Deliberation Strategy:} In the alternative configuration, the Meta-Jury is prompted to act as an independent appellate judge. Rather than calculating weighted averages, it is instructed to qualitatively evaluate the juror justifications and select the severity score best supported by the historical evidence, regardless of how many jurors hold that position. Human review is flagged based on the Meta-Jury's qualitative assessment of the disagreement rather than a strict numerical threshold.
\end{enumerate}

For example, if four agents assign severity 2 with low confidence
and one assigns severity 5 with detailed reasoning citing specific
historiographical omissions, the Heuristic strategy produces a
confidence-weighted average near 2, while the Independent strategy
may select severity 5 as better supported by the evidence.

Both prompting strategies require the Meta-Jury to output a standardized JSON object containing the final severity score, the finalized taxonomy category, a synthesized justification summary, and a boolean flag indicating if human intervention is required. This dual-prompt approach allows us to observe whether strict algorithmic consensus or independent LLM judgment yields better results in human evaluations. 

\section{Experimental Setup}
\label{sec:experiments}

\subsection{Dataset}
\label{sec:dataset}

We constructed an evaluation corpus from history textbooks certified by the Romanian Ministry of Education, sourced from major publishers and covering the full upper-secondary curriculum. The selection of this corpus is motivated by three strategic considerations:
\begin{itemize}
    \item \textbf{Reproducibility and Open Access:} The textbooks are publicly available via the ministry's official digital repository,\footnote{\url{https://www.manuale.edu.ro}} allowing for full study reproducibility by independent researchers without restrictive institutional access agreements.
    
    \item \textbf{Historiographical Domain Competence:} The authors possess direct familiarity with Romanian secondary historiography and its contested periods, including Romania's role in World War II, the Holocaust, and the Communist era, enabling meaningful qualitative validation of pipeline outputs and ensuring findings are historiographically grounded.
    
    \item \textbf{Policy Relevance and Deployment Urgency:} Romania is currently undergoing a comprehensive national curriculum reform~\cite{romania2025curriculum}. Given the massive volume of new materials introduced simultaneously, our agentic auditing architecture serves as a decision-support tool, helping specialized committees rapidly surface pedagogical distortions before final approval.
\end{itemize}

The corpus also presents three technical challenges that directly motivate our architectural design and stress-test its robustness:
\begin{itemize}
    \item \textbf{Content Sensitivity:} The curriculum requires a fine-grained distinction between factual recounting and ideological framing in sensitive historical periods, where standard models often suffer from ``safety-alignment inflation'' and over-penalize factual descriptions~\cite{rottger2023xstest}.
    
    \item \textbf{Multimodal Complexity:} Textbooks utilize non-linear layouts (sidebars, primary sources, captions) where essential context is distributed spatially, necessitating our vision-based Stage 1 screening analysis rather than simple linear text extraction.
    
    \item \textbf{Cultural Alignment Gap:} Safety-aligned models are predominantly calibrated against Western educational norms~\cite{rottger2023xstest,liang2021holistic}. A heterogeneous Stage 2 jury spanning North American, European, and East Asian training distributions explicitly mitigates the influence of any single model's cultural alignment profile.
\end{itemize}

\subsection{Model Selection}
\label{sec:models}

The \textbf{Screening Agent} is implemented using \textit{Llama-4-\-Maverick-\-17B-\-128E-\-Instruct-\-FP8}~\cite{llama4}, selected for its native multimodal capabilities and ability to process high-resolution page images without OCR preprocessing~\cite{xu2020layoutlm,bai2023qwenvl}.

The \textbf{Jury} comprises five evaluative agents selected for their complementary capability profiles: (1) \textit{Mixtral-8x7B-Instruct-v0.1}, a sparse mixture-of-experts model providing multilingual European training coverage relevant to the Romanian corpus~\cite{jiang2024mixtral}; (2) \textit{GPT-OSS-120B}, a dense generalist model utilized for its broad historical knowledge and stable baseline performance~\cite{openai2025gptoss}; (3) \textit{DeepSeek-V3.1}, a reasoning-optimized model chosen for high-fidelity structured attribution and framing analysis~\cite{deepseekv31}; (4) \textit{Cogito-v2-1-671B}, an ultra-large-scale model (671B) included to capture historiographical nuances that may be lost at smaller parameter counts~\cite{cogito2025}; and (5) \textit{Kimi-K2-Thinking}, an extended-reasoning model that externalizes deliberation via inspectable reasoning traces, ensuring the attribution logic is auditable~\cite{kimi2025thinking}.

\textbf{Verdict Synthesis} (Stage 3, Section~\ref{sec:meta}) is performed by a \textit{GPT-5.2} meta-agent~\cite{openai2026gpt52}. This model adjudicates the jury's independent outputs—including category, severity, confidence, and reasoning—to produce a final verdict under the two experimental aggregation configurations.

\subsection{Implementation Details and Report Generation}
\label{sec:implementation}

\paragraph{Data Processing and Parsing.} The input documents are segmented into non-overlapping five-page batches, rendered at 200 DPI (max 1280px). All outputs are serialized as structured JSON, with a three-attempt retry logic for schema validation. Failed parses are discarded to maintain pipeline integrity.

\paragraph{Token Allocation and Cost.} Extended-reasoning models (Kimi-K2-Thinking, DeepSeek-V3.1) are allocated 16,000 output tokens; standard models are capped at 4,096. Processing a full textbook costs approximately \$2, with the jury stage dominating ($\sim$70\%, of which Kimi-K2-Thinking alone accounts for $\sim$50\%), followed by the meta-agent ($\sim$20\%) and screening ($\sim$10\%). All models were accessed via third-party inference APIs,\footnote{\url{https://www.together.ai/}} with the exception of the meta-agent, which was accessed directly via the official OpenAI API.

\paragraph{Stochastic Variation.} Due to the non-deterministic nature of the screening agent, the two evaluated configurations operated on slightly different candidate sets (Heuristic: 281; Independent: 270). While qualitative patterns remained consistent, cross-configuration severity statistics are presented as indicative, as the meta-agent strategies were evaluated on these naturally occurring variations rather than an identical frozen set.

\subsection{Human Evaluation Methodology}
To assess practical utility, we conducted a blind comparison of anonymized reports from the three configurations: Zero-shot, Heuristic, and Independent Deliberation. For each textbook, evaluators selected the most accurate and pedagogically sound report via a public interface,\footnote{\url{https://submission-its.github.io/bias-detection/}} justifying their choice through criteria such as analysis depth, taxonomy application, and tone objectivity.

\begin{figure}[H]
     \centering
     \includegraphics[width=0.73\textwidth]{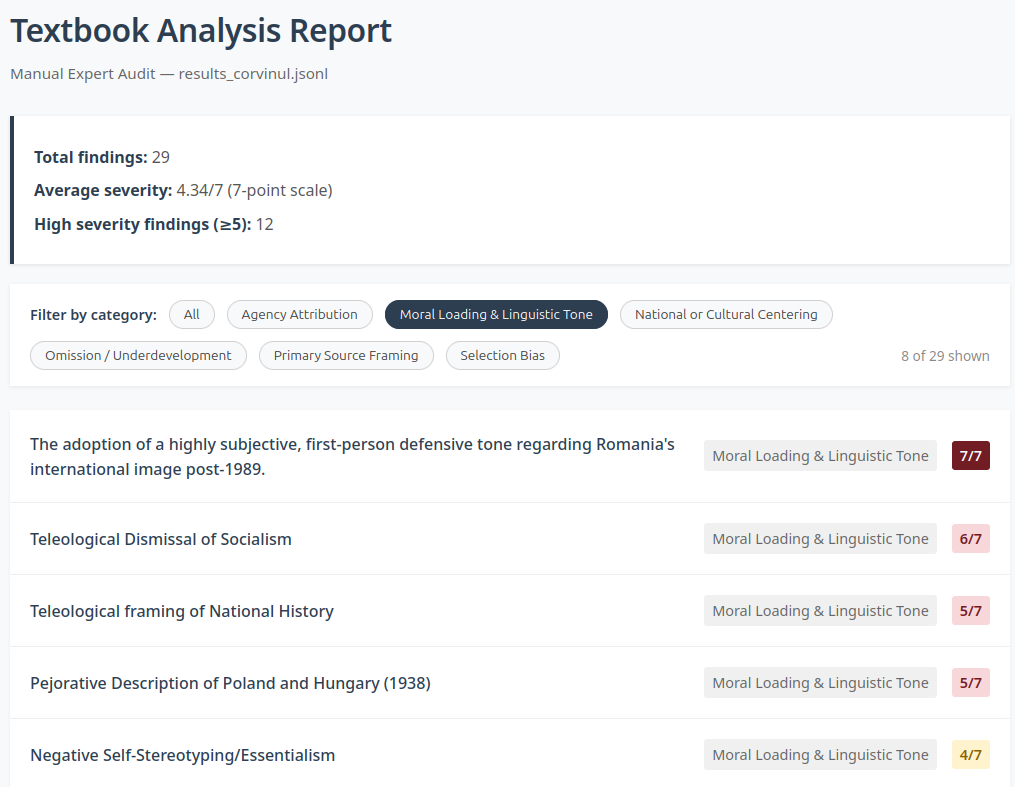}
     \caption{A generated HTML report highlighting extracted historical bias, taxonomy categorization, and assigned severity scores}
     \label{fig:report_ui}
\end{figure}

\textbf{Report Format.}
Pipeline outputs are serialized as static HTML reports (see Fig.~\ref{fig:report_ui}), presenting
each flagged excerpt alongside its attribution label, taxonomy
category, severity score, and the synthesized jury reasoning. 
The full corpus of reports and underlying JSON data was anonymized 
and published to a public GitHub repository to enable evaluator 
access and independent reproducibility.

\section{Results and Discussion}
\label{sec:results}

\subsection{Severity Calibration: Agentic Pipeline vs.\ Zero-Shot Baseline}
\label{sec:severity_results}

The screening agent flagged 270 candidate excerpts, of which 225 (83.3\%) 
received a final severity of 3 or below following jury adjudication---
classified as pedagogically acceptable. Of the remaining 45 cases with 
severity $\geq 4$, the most common categories were \textit{Selection Bias} 
and \textit{Omission / Underdevelopment}. Table~\ref{tab:severity_dist} 
presents the full distribution.
\begin{table}[H]
\centering
\setlength{\tabcolsep}{4pt}
\caption{Distribution of final severity scores ($N=270$,
Independent Deliberation configuration)}
\label{tab:severity_dist}
\small
\begin{tabular}{clrr}
\hline
\textbf{Severity} & \textbf{Label} & \textbf{Count} & \textbf{\%} \\
\hline
1 & Neutral / Pedagogically Sound & 6 & 2.2 \\
2 & Negligible Framing & 63 & 23.3 \\
3 & Minor Imbalance & 156 & 57.8 \\
4 & Moderate Bias & 43 & 15.9 \\
5 & Significant Distortion & 2 & 0.7 \\
6--7 & Severe / Educational Harm & 0 & 0.0 \\
\hline
\end{tabular}
\end{table}

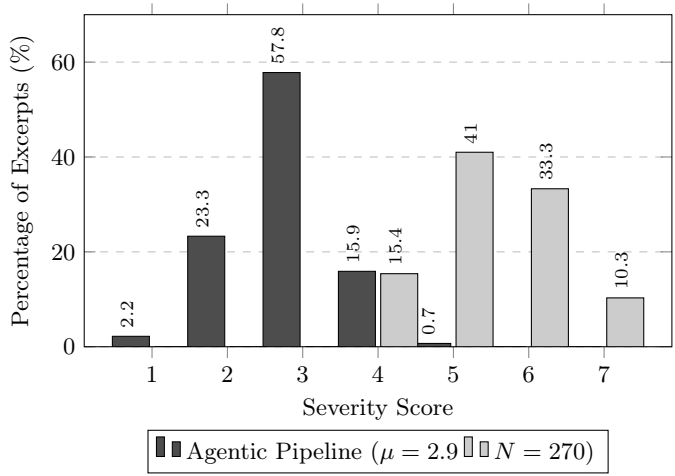
\begin{figure}[H]
    \centering
    \begin{tikzpicture}
    \begin{axis}[
        ybar,
        bar width=14pt, 
        width=0.77\textwidth, 
        height=6cm, 
        enlarge x limits=0.15,
        legend style={at={(0.5,-0.25)}, anchor=north, legend columns=2, font=\footnotesize},
        ylabel={Percentage of Excerpts (\%)},
        xlabel={Severity Score},
        symbolic x coords={1, 2, 3, 4, 5, 6, 7},
        xtick=data,
        ymin=0, ymax=70, 
        ymajorgrids=true,
        grid style=dashed,
        nodes near coords={%
            \pgfmathfloatifflags{\pgfplotspointmeta}{0}{}{\pgfmathprintnumber{\pgfplotspointmeta}}%
        },
        every node near coord/.append style={
            font=\scriptsize,
            rotate=90,
            anchor=west,
            yshift=0.5mm 
        },
    ]
    
    \addplot[fill=black!70, draw=black] coordinates
        {(1,2.2) (2,23.3) (3,57.8) (4,15.9) (5,0.7) (6,0) (7,0)};
        
    \addplot[fill=black!20, draw=black] coordinates
        {(1,0) (2,0) (3,0) (4,15.4) (5,41.0) (6,33.3) (7,10.3)};
        
    \legend{Agentic Pipeline ($\mu=2.9$, $N=270$),
            Zero-Shot Baseline ($\mu=5.4$, $N=39$)}
    \end{axis}
    \end{tikzpicture}
\caption{Severity distributions for the agentic pipeline (dark gray, $\mu=2.9$) 
and zero-shot baseline (light gray, $\mu=5.4$). Samples differ in size and 
configuration.}
    \label{fig:severity_comparison}
\end{figure}

A zero-shot, single-model baseline auditing the identical full textbooks exhibited severe recall failure, extracting only 39 candidate excerpts compared to the pipeline's 270. Furthermore, it systematically inflated severity scores for these 39 excerpts, assigning a mean of 5.4/7 (with no score below 4). In contrast, the agentic pipeline averaged 2.9/7 across its comprehensive set of 270 candidates (Fig.~\ref{fig:severity_comparison}). While this extraction failure precludes a strict per-excerpt comparison, the yield discrepancy validates the need for a dedicated screening agent, just as the inflated severity highlights the baseline's lack of deliberative nuance.

To isolate the impact of the multi-agent architecture from context window and prompting variables, we conducted an ablation study evaluating the zero-shot baseline on identical 5-page chunks with the explicit leniency instruction. While chunking resolved the baseline's initial extraction failure, it transformed the single model into an over-sensitive filter, yielding 192 total excerpts compared to the deliberative pipeline's 109. Crucially, the multi-agent jury acts as a vital consensus filter: it suppressed negligible stylistic noise (reducing flags of severity 3 and below from 91.1\% to 81.7\% of total findings) while doubling the concentration of actionable, higher-severity issues (severity 4 and above: 17.4\% vs. 8.9\%). Furthermore, the deliberative architecture generated more confident assessments (mean confidence 0.84 vs. 0.78) with substantially deeper historiographical justifications (mean 268 vs. 207 characters per explanation). This confirms that while chunking enables basic extraction, multi-agent deliberation is strictly necessary to prevent alert fatigue and curate a confident, high-signal report for editorial governance.

The distribution is strongly concentrated around severity 3, with a
thin right tail (2 excerpts at severity 5; none at 6 or 7). This
indicates that the agentic architecture is well-calibrated for
deployment: the jury layer avoids both under-penalization and the
over-penalization that would render an automated auditing tool
untrustworthy in an editorial governance workflow.

\paragraph{Inter-Agent Agreement.}
All five jury agents produced valid evaluations for 96.3\% of excerpts
(260/270); the remaining 10 cases retained four valid agents after retry
exhaustion. The mean inter-agent severity range was 1.28 points on the
7-point scale, with 69.6\% of excerpts showing ranges of $\leq 1$
point---indicating strong consensus on the majority of cases. Only
3.3\% exhibited ranges of $\geq 3$ points. The meta-agent escalation
mechanism flagged 18 cases (6.7\%) for human review, demonstrating
selective rather than indiscriminate escalation.

\subsection{Role of Source Attribution in Severity Calibration}
\label{sec:attribution_results}

To assess the effect of attribution classification on downstream jury
behavior, we compared severity outcomes across the two categories
assigned during screening. Excerpts classified as \textit{Primary
Source Usage} ($N=56$) received a mean final severity of 2.68/7,
compared to 2.95/7 for \textit{Textbook Narrative} excerpts ($N=214$).
While the absolute difference is modest (0.27 points), the direction
is consistent with the intended design: jury agents apply more lenient
evaluation to historical quotations than to authorial exposition.

Qualitative examination of jury agent reasoning for primary source
excerpts showed a consistent shift from normative condemnation of
archaic language toward pedagogical evaluation of contextual
framing---a pattern observed across multiple agents in the jury.
These findings suggest that explicit attribution enforcement
meaningfully conditions jury evaluation, reducing the risk of
misclassifying historically situated language as authorial bias.

\subsection{Key Historiographical Bias Patterns}
\label{sec:bias_patterns}

While overt hate speech was absent from the evaluated corpus, the
pipeline identified prevalent structural biases of direct relevance
to pre-publication governance review. The most frequent and severe
failure mode was historical sanitization through selective omission:
accounts of Romania's entry into World War II frequently framed
military participation strictly as territorial recovery, omitting
ideological alignment with Nazi Germany and state-sponsored mass
violence. Jury agents assigned the highest severity scores (up to
5/7) to these excerpts, classifying them as substantive distortions
rather than neutral summarizations.

Beyond omission, the architecture consistently detected ethno-nationalist
essentialism and Eurocentric framing. Textbooks frequently presented
ethnogenesis narratives as biological or metaphysical continuities,
suppressing alternative historiographical debates and framing national
identity as a fixed, inherent attribute. Excerpts classified under
\textit{Perspective Limitation} produced the highest category-level
mean severity at 3.4/7 ($N=15$), followed by \textit{Omission /
Underdevelopment} at 3.08/7 ($N=26$). Table~\ref{tab:category_dist}
summarizes the most frequent taxonomy categories.

\begin{table}[H]
\centering
\caption{Most frequent bias taxonomy categories with mean severity
scores (Independent Deliberation, $N=270$)}
\label{tab:category_dist}
\begin{tabular}{lrr}
\hline
\textbf{Category} & \textbf{Count} & \textbf{Mean Severity} \\
\hline
Narrative Framing & 68 & 2.88 \\
Primary Source Framing & 47 & 2.62 \\
Selection Bias & 38 & 3.05 \\
Omission / Underdevelopment & 26 & 3.08 \\
Moral Loading & 23 & 2.83 \\
National or Cultural Centering & 16 & 2.88 \\
Perspective Limitation & 15 & 3.40 \\
Source Selection Bias & 12 & 2.92 \\
Teleological Narrative & 8 & 3.12 \\
Other (4 categories) & 17 & 2.71 \\
\hline
\end{tabular}
\end{table}

Notably, while the surface-level sentiment of cultural descriptions
often appeared positive or neutral, the reasoning-oriented agents within
the jury exposed unexamined normative assumptions beneath the text.
These patterns, structural omissions and essentialist framing rather
than overt hate speech, are precisely the type of subtle
historiographical distortion that is difficult to detect manually at
scale and that a pre-publication governance tool must surface reliably.

\subsection{Human Evaluation of Report Quality}
\label{sec:human_eval_results}

We conducted a blind comparative study with 18 evaluators (history and computer science students, University of Bucharest) performing 54 paired comparisons across three textbooks. Evaluators were provided with the source material alongside three anonymized HTML reports, one per configuration, and selected the most accurate, comprehensive, and pedagogically sound relative to the original text. Report assignments were randomized to prevent positional bias.

\begin{figure}[H]
    \centering
    \begin{tikzpicture}
    \begin{axis}[
        ybar,
        bar width=14pt,
        width=0.9\textwidth,
        height=5cm,
        enlarge x limits=0.25,
        legend style={at={(0.5,-0.30)}, anchor=north, legend columns=1,
        font=\footnotesize},
        ylabel={Number of Votes (out of 18)},
        symbolic x coords={Corvinul (11th), Niculescu (12th),
        Sigma (11th)},
        xtick=data,
        nodes near coords,
        nodes near coords align={vertical},
        ymin=0, ymax=15,
        ymajorgrids=true,
        grid style=dashed,
    ]
    \addplot[fill=black!80, draw=black]
        coordinates {(Corvinul (11th),11) (Niculescu (12th),12)
        (Sigma (11th),12)};
        
    \addplot[fill=black!40, draw=black]
        coordinates {(Corvinul (11th),6) (Niculescu (12th),4)
        (Sigma (11th),4)};
        
    \addplot[fill=black!10, draw=black]
        coordinates {(Corvinul (11th),1) (Niculescu (12th),2)
        (Sigma (11th),2)};
        
    \legend{Independent Deliberation, Heuristic Aggregation,
            Zero-Shot Baseline}
    \end{axis}
    \end{tikzpicture}
   \caption{Evaluator preferences ($N=54$ comparisons). Independent Deliberation was consistently preferred in randomized blind testing across all textbooks.}
    \label{fig:evaluation_results}
\end{figure}
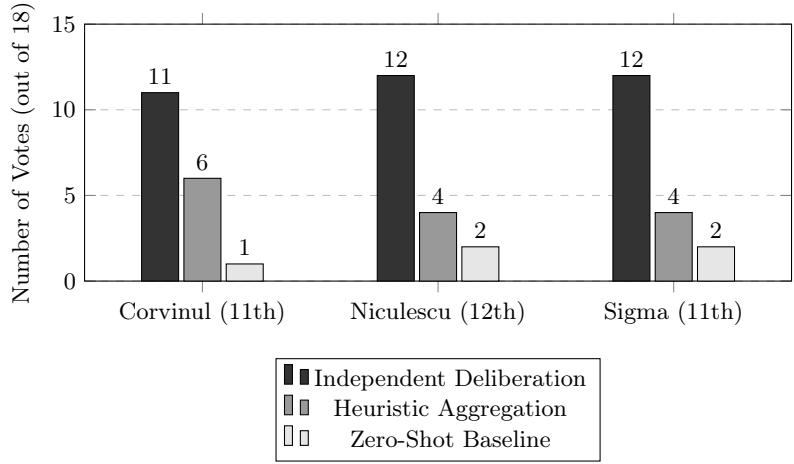

The Independent Deliberation configuration was preferred in 64.8\% of 
cases (35/54), followed by Heuristic Aggregation (25.9\%) and the 
Zero-Shot Baseline (9.3\%), with consistent rankings across all three 
textbooks and publishers (Fig.~\ref{fig:evaluation_results}). The 
preference for Independent Deliberation suggests that qualitative jury 
synthesis---particularly surfacing well-supported minority positions---
yields more actionable reports than fixed numerical thresholds. The 
near-total rejection of the Zero-Shot Baseline confirms that single-pass 
auditing is insufficient for educational governance.

The most cited selection criteria were Clarity and structure (50/54 
evaluations), Correct identification of issues (41/54), Depth of 
analysis (35/54), and Comprehensiveness (33/54), confirming that 
readability is a primary determinant of usability. As this evaluation 
measured report quality rather than per-excerpt detection accuracy, 
establishing formal precision and recall against expert-annotated ground 
truth remains a priority for future work.
\subsection{Limitations and Future Work}
\label{sec:limitations}

Several limitations constrain our findings. First, our comparative analysis between the agentic pipeline and the zero-shot baseline illustrates the delta between a specialized architecture and a naive deployment scenario, which introduces significant confounding variables. The baseline processed entire textbooks within a single context window, precipitating severe recall failure ($N=39$ vs. $270$), whereas the screening agent processed localized five-page batches. This discrepancy precludes a controlled per-excerpt comparison. Additionally, the reliance on a single model in Stage 1 introduces a recall risk: excerpts missed by the screener are never surfaced for jury evaluation. Exploring a multi-model ensemble for screening is a priority for future work. Second, the jury's explicit leniency calibration instruction is absent from the baseline. Consequently, no current ablation isolates whether the observed severity reduction stems independently from the multi-agent architecture, the Source Attribution Schema, or simply the prompt-level calibration and narrowed context window. Finally, our human evaluation assessed report usability rather than detection accuracy; establishing formal precision and recall requires expert-annotated ground truth.

Future work must address these gaps through controlled ablation studies, such as evaluating a single-model baseline on identical five-page chunks with the exact leniency prompt, and rigorous expert annotation. Beyond governance, the structured controversy reports suggest a student-facing direction: a tool allowing students to interrogate contested passages in real time, extending the system from pre-publication auditing toward supporting critical historical thinking in the classroom.

\section{Conclusions}
\label{sec:conclusion}

This paper introduces an automated auditing architecture for detecting subtle historiographical biases in high-stakes educational materials. By decomposing the task into broad-coverage discovery and deliberative adjudication, we address two primary failure modes of previous approaches: the inability to detect structural omission and the tendency to hallucinate bias in primary sources.

The system identified significant patterns of ethno-nationalist essentialism and sanitization by omission across the evaluated corpus, validating its capacity to surface meaningful historiographical concerns rather than superficial flags. The Meta-Jury selectively escalated only 6.7\% of cases for human review, demonstrating that the architecture can prioritize expert attention without overwhelming reviewers---a property essential for practical deployment.

Beyond methodological contributions, this work addresses a real-world governance need. Romania's ongoing curriculum reform requires verifying a massive volume of new history textbooks simultaneously---making exhaustive manual bias analysis infeasible. At approximately \$2 per textbook, the architecture offers a scalable decision-support tool for review committees, surfacing controversies before materials reach classrooms without replacing expert judgment. Future work will focus on cross-national corpus replication, controlled ablation studies, expert-annotated ground truth, and a student-facing interface for in-classroom critical inquiry.

\bibliographystyle{splncs04}
\bibliography{main}

@book{apple2014official,
author = {Apple, Michael},
year = {2014},
pages = {1-233},
title = {Official knowledge: Democratic education in a conservative age, third edition},
publisher = {Routledge},
isbn = {9781136706806},
doi = {10.4324/9780203814383}
}

@book{pingel2010unesco,
  title={UNESCO Guide on Textbook Research and Textbook Revision},
  author={Pingel, Falk},
  year={2010},
  publisher={UNESCO},
  address={Paris}
}

@article{kasneci2023chatgpt,
  title={{ChatGPT} for good? On opportunities and challenges of large language models for education},
  author={Kasneci, Enkelejda and Sessler, Kathrin and K{\"u}chemann, Stefan and others},
  journal={Learning and Individual Differences},
  volume={103},
  pages={102274},
  year={2023},
  publisher={Elsevier},
  doi={10.1016/j.lindif.2023.102274}
}

@inproceedings{liang2021holistic,
  title={Holistic Evaluation of Language Models},
  author={Liang, Percy and Bommasani, Rishi and Lee, Tony and others},
  booktitle={TMLR},
  year={2023},
  doi={10.48550/arXiv.2211.09110}

}

@article{du2023improving,
  title={Improving Factuality and Reasoning in Language Models through Multiagent Debate},
  author={Du, Yilun and Li, Shuang and Torralba, Antonio and Tenenbaum and others},
  journal={arXiv preprint arXiv:2305.14325},
  year={2023},
  url={https://arxiv.org/abs/2305.14325}
}

@inproceedings{lucy2020content,
  title={Content Analysis of Textbooks via Natural Language Processing: Findings on Gender, Race, and Ethnicity in Texas US History Textbooks},
  author={Lucy, Li and Demszky, Dorottya and Bromley, Patricia and Jurafsky, Dan},
  booktitle={AERA Open},
  year={2020},
  doi = {10.1177/2332858420940312}
}

@article{ji2023survey,
  title={Survey of hallucination in natural language generation},
  author={Ji, Ziwei and Lee, Nayeon and Frieske, Rita and others},
  journal={ACM Computing Surveys},
  volume={55},
  number={12},
  pages={1--38},
  year={2023},
  doi={10.1145/3571730},
  publisher={ACM New York, NY}
}

@inproceedings{xu2020layoutlm,
  title={LayoutLM: Pre-training of Text and Layout for Document Image Understanding},
  author={Xu, Yiheng and Li, Minghao and Cui, Lei and others},
  booktitle={ACM SIGKDD},
  year={2020},
  doi = {10.1145/3394486.3403172}
}

@misc{bai2023qwenvl,
      title={Qwen-VL: A Versatile Vision-Language Model for Understanding, Localization, Text Reading, and Beyond}, 
      author={Jinze Bai and Shuai Bai and Shusheng Yang and others},
      year={2023},
      doi={10.48550/arXiv.2308.12966}
}

@inproceedings{lewis2020retrieval,
  title={Retrieval-Augmented Generation for Knowledge-Intensive NLP Tasks},
  author={Lewis, Patrick and Perez, Ethan and Piktus, Aleksandra and others},
  booktitle={NeurIPS},
  year={2020},
  doi={10.48550/arXiv.2005.11401}
}

@article{zhang2025siren,
    author = {Zhang, Yue and Li, Yafu and Cui, Leyang and others},
    title = {Siren’s Song in the AI Ocean: A Survey on Hallucination in Large Language Models},
    journal = {Computational Linguistics},
    year = {2025},
    doi = {10.1162/COLI.a.16},
}

@inproceedings{zheng2024judging,
  title={Judging {LLM}-as-a-Judge with {MT-Bench} and Chatbot Arena},
  author={Zheng, Lianmin and Chiang, Wei-Lin and Sheng, Ying and Zhuang and others},
  booktitle={Advances in Neural Information Processing Systems (NeurIPS)},
  volume={36},
  year={2023},
  url={https://arxiv.org/abs/2306.05685}
}

@book{luckin2016intelligence,
  title={Intelligence Unleashed: An Argument for AI in Education},
  author={Luckin, Rose and Holmes, Wayne and Griffiths, Mark and Forcier, Laurie B},
  year={2016},
  publisher={Pearson London},
  ISBN = {9780992424886}
}

@article{zhai2021review,
  title={A Review of Artificial Intelligence (AI) in Education from 2010 to 2020},
  author={Zhai, Xinyou and Chu, Xiaojing and Chai, Ching Sing and others},
  journal={Complexity},
  volume={2021},
  pages={1--18},
  year={2021},
  publisher={Hindawi},
  doi = {10.1155/2021/8812542}
}

@inproceedings{liu2023gevalnlgevaluationusing,
  title={{G-Eval}: {NLG} Evaluation using {GPT-4} with Better Human Alignment},
  author={Liu, Yang and Iter, Dan and Xu, Yichong and others},
  booktitle={Proceedings of EMNLP},
  pages={2511--2522},
  year={2023},
  doi={10.18653/v1/2023.emnlp-main.153}
}

@inproceedings{rottger2023xstest,
  title={XSTest: A Test Suite for Identifying Exaggerated Safety Behaviours in Large Language Models},
  author={R{\"o}ttger, Paul and Kirk, Hannah Rose and Vidgen, Bertie and others},
  booktitle={NAACL},
  year={2024},
  doi ={10.48550/arXiv.2308.01263}
}

@misc{jiang2024mixtral,
      title={Mixtral of Experts}, 
      author={Jiang, Albert Q. and Sablayrolles, Alexandre and Roux, Arthur and others},
      year={2024},
      doi = {10.48550/arXiv.2401.04088}
}

@misc{openai2025gptoss,
  title={gpt-oss-120b \& gpt-oss-20b Model Card},
  author={{OpenAI} and Agarwal, Sandhini and Ahmad, Lama and Ai, Jason and Altman, Sam and others},
  year={2025},
  doi={10.48550/arXiv.2508.10925}
}

@misc{llama4,
  title={Llama 4 Maverick},
  author={{Meta AI} and others},
  year={2025},
  howpublished={\url{https://ai.meta.com/llama/}}
}

@misc{deepseekv31,
  title={DeepSeek-V3 Technical Report},
  author={{DeepSeek-AI} and others},
  year={2024},
  doi={10.48550/arXiv.2412.19437}
}

@misc{cogito2025,
  title={Cogito v2.1 671B Model Card},
  author={{Deep Cogito}},
  year={2025},
  howpublished={\url{https://huggingface.co/deepcogito}}
}

@misc{kimi2025thinking,
  title={Kimi K2: Open Agentic Intelligence},
  author={{Kimi Team} and Bai, Yifan and others},
  year={2025},
  doi={10.48550/arXiv.2507.20534}
}

@misc{openai2026gpt52,
  title={GPT-5.2 Technical Specifications},
  author={{OpenAI}},
  year={2026},
  howpublished={\url{https://developers.openai.com/api/docs/models/gpt-5.2}}
}

@misc{romania2025curriculum,
  author = {{Ministerul Educației}},
  title = {Comunicat de presă nr. 109/2025 privind constituirea grupurilor de lucru pentru elaborarea programelor școlare},
  howpublished = {\url{https://www.edu.ro/press_rel_109_2025_grupuri_lucru_programe_scolare_inv_liceal}},
  year = {2025},
  note = {Accessed: 2025-02-24}
}

@inproceedings{chan2023chateval,
  title={ChatEval: Towards Better LLM Evaluations via Multi-Agent Debate},
  author={Chan, Chi-Min and Chen, Weize and Su, Yusheng and others},
  booktitle={ICLR},
  doi = {10.48550/arXiv.2308.07201},
  year={2024}
}

@book{unesco2023guidance,
  title={Guidance for generative AI in education and research},
  author={UNESCO},
  year={2023},
  publisher={UNESCO Publishing},
  address={Paris}
}

@inproceedings{chen2023agentverse,
  title={AgentVerse: Facilitating Multi-Agent Collaboration and Exploring Emergent Behaviors},
  author={Chen, Weize and Su, Yusheng and Zuo and others},
  booktitle={ICLR},
  year={2024},
  doi={10.48550/arXiv.2308.10848}
}

@article{wu2023autogen,
  title={AutoGen: Enabling Next-Gen LLM Applications via Multi-Agent Conversation},
  author={Wu, Qingyun and Bansal, Gagan and Zhang and others},
  journal={arXiv preprint arXiv:2308.08155},
  year={2023},
  doi={10.48550/arXiv.2308.08155}
}

@article{liu2024lost,
  title={Lost in the middle: How language models use long contexts},
  author={Liu, Nelson F and Lin, Kevin and Chen, Jiajun and others},
  journal={Transactions of the Association for Computational Linguistics},
  volume={12},
  pages={157--173},
  year={2024},
  publisher={MIT Press},
  doi={10.1162/tacl_a_00638}
}

@inproceedings{hsieh2024ruler,
  title={RULER: What's the Real Context Size of Your Long-Context Language Models?},
  author={Hsieh, Cheng-Ping and Simig, Daniel and others},
  booktitle={Proceedings of EMNLP},
  year={2024},
  url={https://arxiv.org/abs/2404.06654}
}

@article{joshi2015,
  author    = {Joshi, Ankur and Kale, Saket and Chandel and others},
  title     = {Likert Scale: Explored and Explained},
  journal   = {British Journal of Applied Science \& Technology},
  volume    = {7},
  number    = {4},
  pages     = {396--403},
  year      = {2015},
  doi       = {10.9734/BJAST/2015/14975}
}

@article{preston2000,
  author    = {Preston, Carolyn C. and Colman, Andrew M.},
  title     = {Optimal number of response categories in rating scales: 
               Reliability, validity, discriminating power, and 
               respondent preferences},
  journal   = {Acta Psychologica},
  volume    = {104},
  number    = {1},
  pages     = {1--15},
  year      = {2000},
  doi       = {10.1016/S0001-6918(99)00050-5}
}

@book{stradling2003,
  author    = {Stradling, Robert},
  title     = {Multiperspectivity in History Teaching: A Guide for Teachers},
  publisher = {Council of Europe Publishing},
  year      = {2003}
}

\end{document}